\documentclass[twoside,leqno,twocolumn]{article}
\usepackage{ltexpprt} 
\usepackage{subfigure}
\usepackage{graphicx}
\usepackage{amsmath}
\usepackage{bm}
\usepackage{url}
\usepackage{stmaryrd}
\usepackage{balance}
\usepackage{amssymb}
\usepackage{booktabs}
\usepackage[latin1]{inputenc}
\usepackage{multirow}

\newcommand{\ie}[0]{\textit{i.e.}}
\newcommand{\ourmethod}[0]{$\text{DuSK}_{\text{RBF}}$}
\newcommand{\ours}[0]{$\text{DuSK}$}

\newtheorem{defn}{\textsc{Definition}}

\begin{document}

\title{{\ours}: A Dual Structure-preserving Kernel for Supervised Tensor Learning 
with Applications to Neuroimages}

\author{Lifang He\thanks{Computer Science and Engineering, South China University of Technology, China. lifanghescut@gmail.com}
\and
Xiangnan Kong\thanks{Computer Science Department, University of Illinois at Chicago, USA. xkong4@uic.edu}
\and
Philip S. Yu \thanks{Computer Science Department, University of Illinois at Chicago, USA. psyu@cs.uic.edu}
\and
Ann B. Ragin \thanks{Department of Radiology, Northwestern University, USA. ann-ragin@northwestern.edu}
\and
Zhifeng Hao \thanks{Faculty of Computer, Guangdong University of Technology, China. mazfhao@scut.edu.cn}
\and
Xiaowei Yang \thanks{School of Sciences, South China University of Technology, China. xwyang@scut.edu.cn}
}

\date{}
\maketitle
\begin{abstract}
With advances in data collection technologies, tensor data is assuming increasing prominence in many applications and the problem of supervised tensor learning has emerged as a topic of critical significance in the data mining and machine learning community. Conventional methods for supervised tensor learning mainly focus on learning kernels by flattening the tensor into vectors or matrices, however structural information within the tensors will be lost. In this paper, we introduce a new scheme to design structure-preserving kernels for supervised tensor learning. Specifically, we demonstrate how to leverage the naturally available structure within the tensorial representation to encode prior knowledge in the kernel. We proposed a tensor kernel that can preserve tensor structures based upon dual-tensorial mapping. The dual-tensorial mapping function can map each tensor instance in the input space to another tensor in the feature space while preserving the tensorial structure. Theoretically, our approach is an extension of the conventional kernels in the vector space to tensor space. We applied our novel kernel in conjunction with SVM to real-world tensor classification problems including brain fMRI classification for three different diseases ({\ie}, Alzheimer's disease, ADHD and brain damage by HIV). Extensive empirical studies demonstrate that our proposed approach can effectively boost tensor classification performances, particularly with small sample sizes.

\end{abstract}

\section{Introduction}
Supervised learning is one of the most fundamental data mining tasks. Conventional approaches on supervised learning usually assume, explicitly or implicitly, that data instances are represented as feature vectors. However, in many real-world applications, data instances are more naturally represented as second-order (matrices) or higher-order tensors, where the order of a tensor corresponds to the number of modes or ways. 
For example, in computer vision, a grey-level image is inherently a 2-D object, which can be represented as a second-order tensor with the column and row modes \cite{YXYZTZ07}; in medical neuroimaging, an MRI (Magnetic Resonance Imaging) image is naturally a third-order tensor consisting of 3-D voxels \cite{A13}. Supervised learning on this type of data is called supervised tensor learning, where each instance in the input space is represented as a tensor. With the rapid proliferation of tensor data, supervised tensor learning has drawn significant attention in recent years in the machine learning and data mining communities.

A straightforward solution to supervised tensor learning is to convert the input tensors into feature vectors, and feed the feature vectors to a conventional supervised learning algorithm. However, tensor objects are commonly specified in high-dimensional space. For example, a typical MRI image of size $256\times 256\times 256$ voxels contains $16,777,216$ features \cite{ZLZ12}. This makes traditional methods prone to overfitting, especially for small sample size problems \cite{DD08}. On the other hand, tensorial representations retain the information about the structure of the high-dimensional space the data lie in, such as about the spatial arrangement of the voxel-based features in a 3-D image. When converting tensors into vectors, such important structural information will be lost. In particular, the entries of a tensor object are often highly correlated with surrounding entries. For example, in MRI image data, adjacent voxels usually exhibit similar patterns, which means that the source images contain redundant information at this voxel. It is believed by many researchers that potentially more compact and useful representations can be extracted from the original tensor data and thus result in more accurate and interpretable models. Therefore, supervised learning algorithms operating directly on tensors rather than their vectorized versions are much desired.

\begin{figure}[t]
\centering
\subfigure[Vector-based kernels]
{
    \label{fig1_vector_a}
    \begin{minipage}{.8\columnwidth}
        \centering
        \includegraphics[width=6.0cm]{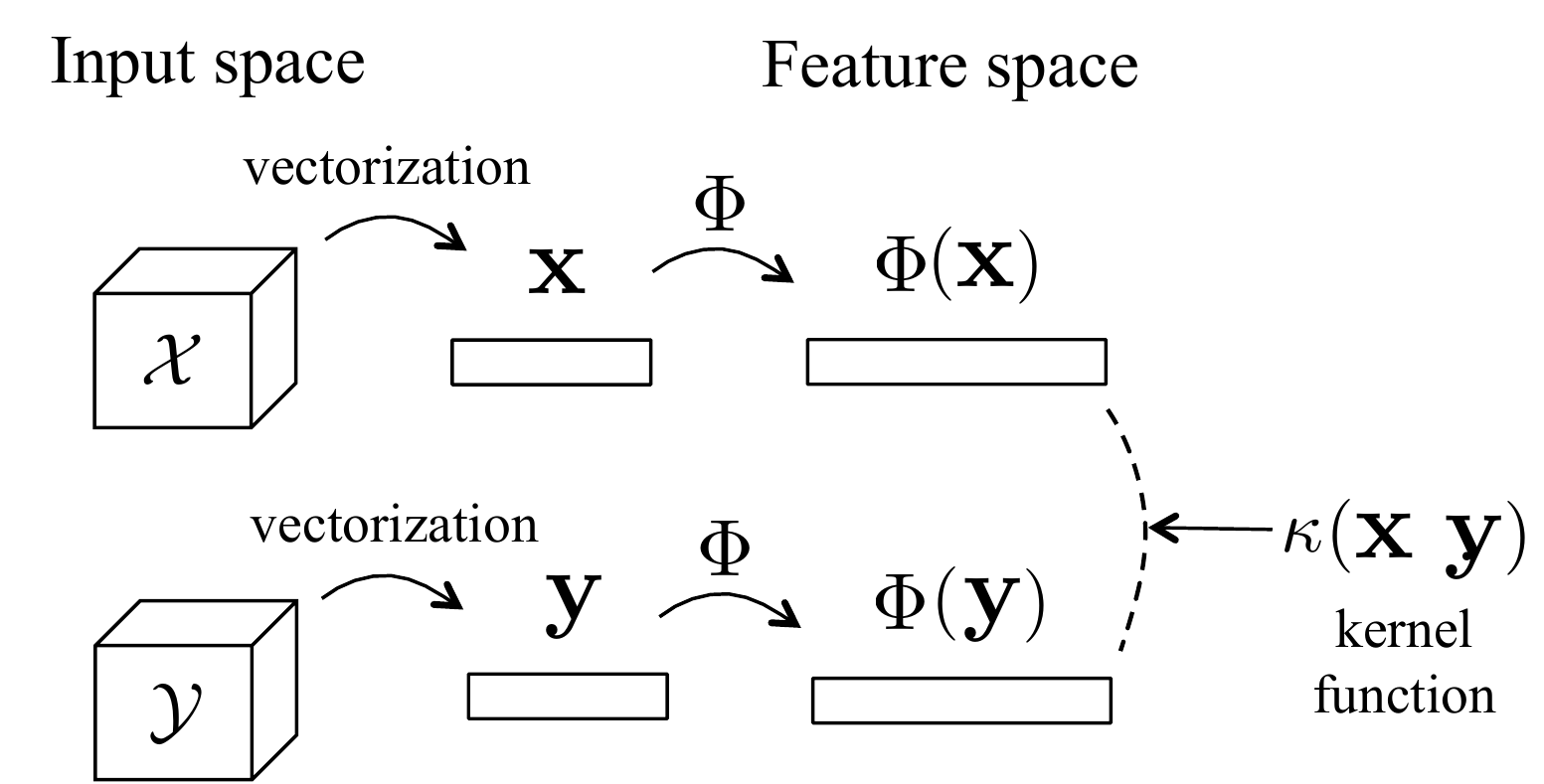}
    \end{minipage}
}
\subfigure[Conventional tensor kernels]
{
    \label{fig1_matrix_b}
    \begin{minipage}{.7\columnwidth}
        \centering
        \includegraphics[width=7.0cm]{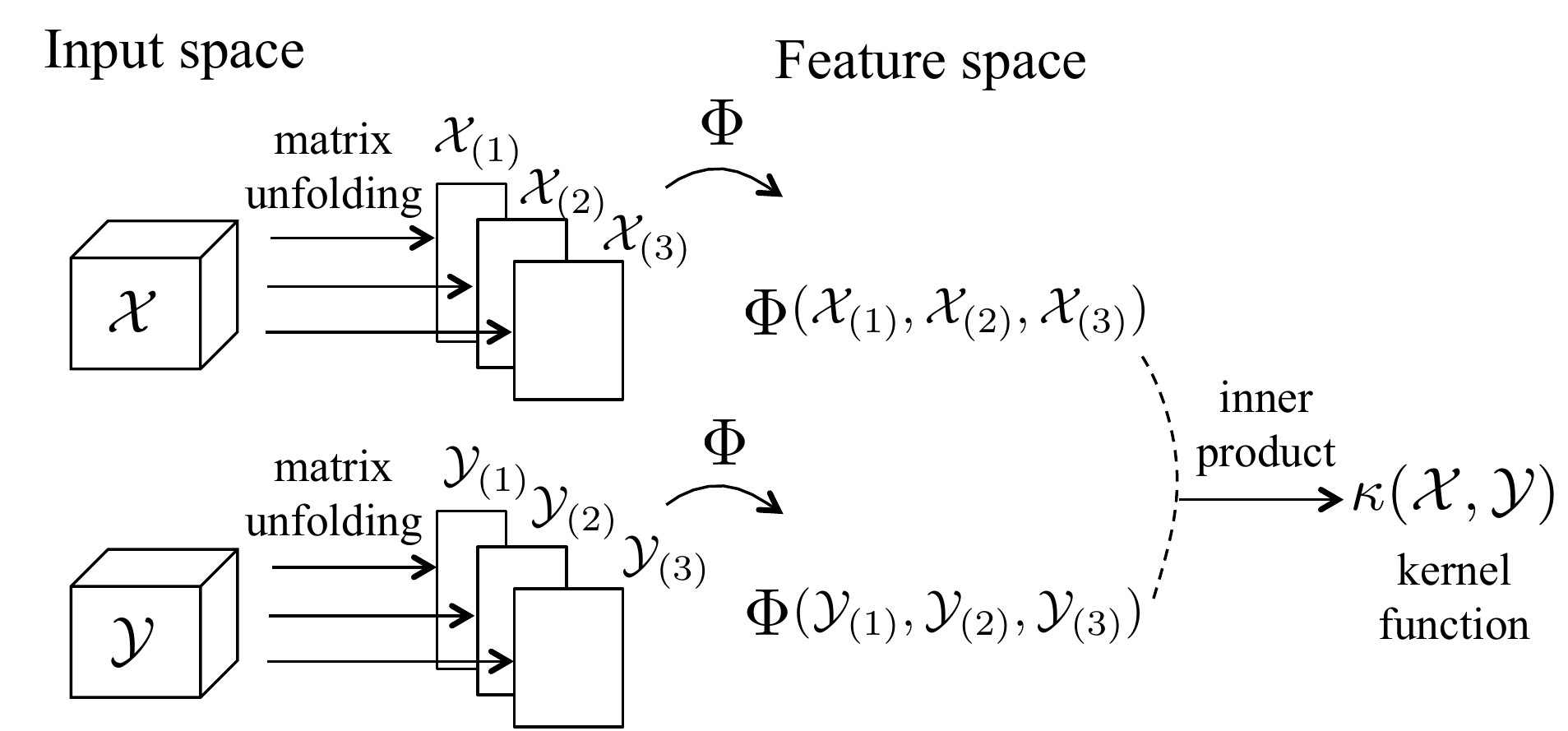}
    \end{minipage}
}
\subfigure[Our {\ours}]
{
    \label{fig1_tensor_c}
    \begin{minipage}{.8\columnwidth}
        \centering
        \includegraphics[width=6.0cm]{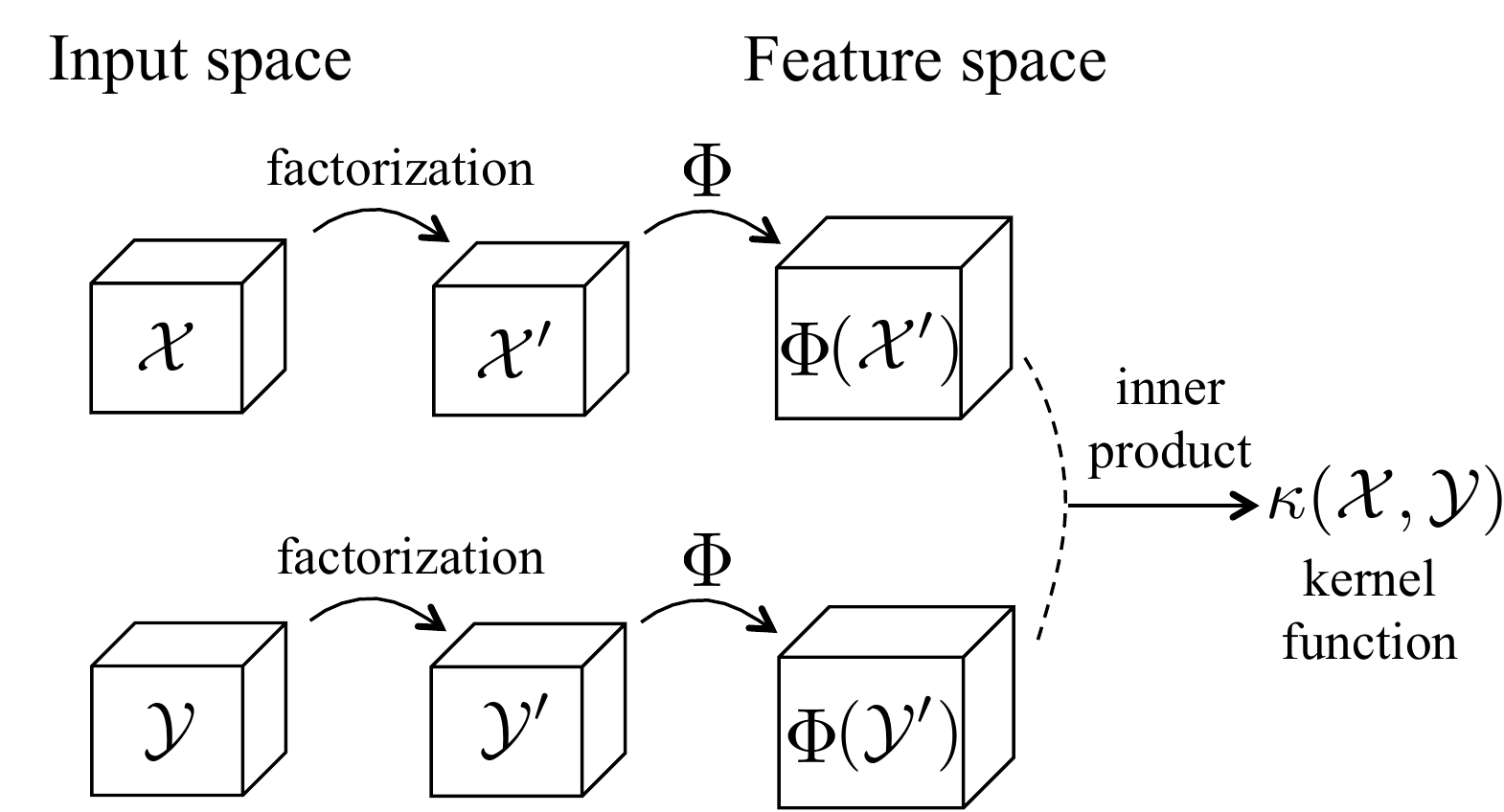}
    \end{minipage}
}\vspace{-8pt}
\caption{Schematic view of the key difference among three kernel learning schemes. Standard kernel (a) works on the vectorized representation and conventional tensor-based kernel (b) applies tensor-to-matrix alignment first, which
may lead to loss of structural information. Our method (c) works on the tensor representation directly.}
\label{fig_example1}
\vspace{-10pt}
\end{figure}

Formally, a major difficulty in supervised tensor learning is how to build predictive models that can leverage the naturally available structure of tensor data to facilitate the learning process. In the literature, several solutions have been proposed. Previous work on supervised tensor learning mainly focuses on linear models \cite{CHH06,GKP12,HHCY13,TLWH07,ZLZ12}, which assume, explicitly or implicitly, that data are linearly separable in the input space. However, in practice this assumption is often violated and the linear decision boundaries do not adequately separate the classes. Recently, several approaches try to exploit the tensor structure with nonlinear kernel models \cite{SLS11,SOLS12,ZZAZC13}, which first unfold the tensor along each of its modes, and then use these unfolded matrices to construct nonlinear kernels for supervised tensor learning as shown in Figure~\ref{fig1_matrix_b}.
However, these methods can only capture the relationships within each single mode of the tensor data, because the structural information about inter-mode relationships of tensor data is lost in the unfolding procedures. 

In this paper, we study the problem of supervised tensor learning with nonlinear kernels which can adequately preserve and utilize the structure of the tensor data. The major research challenges of supervised tensor learning with structure-preserving kernels can be summarized as follows:

\noindent\textbullet \ 
\textbf{High-dimensional tensors}: One fundamental problem in supervised tensor learning lies in the intrinsic high dimensionality of tensor objects. 
Traditional supervised learning algorithms assume that the instances are represented as vectors. However, in the context of tensors, each data object is usually not represented as a vector but a high-dimensional multi-mode (also known as multi-way) array. If we reshape the tensor into a vector, the number of features is extremely high. Both computability and theoretical guarantee of the traditional models are compromised by this ultra-high dimensionality.\\
\noindent\textbullet \ 
\textbf{Complex tensor structure}:
Another fundamental problem in supervised tensor learning lies in complex structure of tensors. 
Conventional tensor-based kernel approaches focus on unfolding tensor data into matrices \cite{SLS11,SOLS12,ZZAZC13} which can only preserve the one-way relationships within the tensor data. 
However, in many real-world applications, the tensor data have multi-way structures. Such prior knowledge about multi-way relationships among features should be incorporated to build more accurate and interpretable models, especially in the case of high dimensional tensor data with small sample size.\\
\noindent\textbullet \ 
\textbf{Nonlinear separability}: In real-world applications, the data is usually not linearly separable in the input space. Conventional supervised tensor learning methods which can preserve tensor structures are often based upon linear models. Thus these methods cannot efficiently solve nonlinear learning problems on tensor data.

In this paper, we propose a novel approach to supervised tensor learning, called {\ours} (Dual Structure-preserving Kernels). Our framework is illustrated in Figure~\ref{fig1_tensor_c}.
Different from conventional methods, our approach is based upon kernel methods and tensor factorization techniques that can fully capture the multi-way structures of tensor data. We first extract a more compact and informative representation from the original data using a tensor factorization method, {\ie}, CANDECOMP/PARAFAC (CP) \cite{KB09}. Then we define a structure-preserving feature mapping to derive the {\ours} kernels in the tensor product feature space, used in conjunction with kernel machines to solve the supervised tensor learning problems. 
Empirical studies on real-world tasks (classifying fMRI images of different brain diseases, {\ie}, Alzheimer's disease, ADHD and HIV) demonstrate that the proposed approach can significantly boost the classification performances on tensor datasets.

\section{PRELIMINARIES} \label{sec_formulation}
Before presenting our approach, we introduce some related concepts and notation of tensors. Table~\ref{tab_notation} lists some basic symbols defined in this study. We first give a formal mathematical definition of the tensor, which provides an intuitive understanding of the algebraic structure of the tensor that tensor object has the tensor product structure.
\begin{defn}[Tensor] An $N$th-order tensor is an element of the tensor product of $N$ vector spaces, each of which has its own coordinate system.
\end{defn}

We use $\mathcal{A}=\left(a_{i_1,i_2,\ldots,i_N}\right) \in \mathbb{R}^{I_{1} \times I_{2} \times \cdots \times I_{N}}$
to denote a tensor $\mathcal{A}$ of $N$ order. For $n=1, 2, \cdots, N$, $I_{n}$ is the dimension of $\mathcal{A}$ along the $n$-th mode.
Based on the above definition, we define inner product, tensor norm, tensor product, and rank of a tensor and give CP model as follows:

\begin{defn}[Inner product] The inner product of two same-sized tensors $\mathcal{A}, \mathcal{B} \in \mathbb{R}^{I_{1} \times I_{2} \times \cdots \times I_{N}}$ is defined as the sum of the products of their entries:
\begin{equation}\label{eq1}
\left\langle \mathcal {A}, \mathcal {B}\right\rangle=\sum_{i_{1}=1}^{I_1}\sum_{i_{2}=1}^{I_2}\cdots\sum_{i_{N}=1}^{I_N}a_{i_1,i_2,\ldots,i_N}b_{i_1,i_2,\ldots,i_N}.
\end{equation}
\end{defn}

\begin{defn}[Tensor norm]
The norm of a tensor $\mathcal {A}$ is defined to be the square root of the sum of all entries of the tensor squared, {\ie},
\begin{equation}\label{eq2}
\left\|\mathcal {A}\right\|_{F}=\sqrt{\left\langle \mathcal {A}, \mathcal {A}\right\rangle}=\sqrt{\sum_{i_1=1}^{I_1}\sum_{i_2=1}^{I_2}\cdots\sum_{i_N=1}^{I_N}a_{i_1,i_2,\ldots,i_N}^{2}}.
\end{equation}
\end{defn}
As we see the norm of a tensor is a straightforward generalization of the usual Frobenius norm for matrices and of the $l_{2}$ norm for vectors.

\begin{table}[t]
\centering
{\scriptsize
\caption{List of symbols}\label{tab_notation}
\begin{tabular}{ll}
\toprule
Symbol & Definition and Description\\
\midrule
$s$ & each lower-case represents a scale\\
$\mathbf{v}$ & each boldface lowercase letter represents a vector\\
$\mathbf{M}$ & each boldface capital letter represents a matrix\\
$\mathcal{T}$ & each calligraphic letter represents a tensor\\
$\mathfrak{G}$ & each gothic letter represent a general set or space\\
$\otimes$ & denotes tensor product\\
$\left\langle . ,. \right\rangle$ & denotes the inner product in some feature space\\
$R=$\textit{Rank}($\mathcal{A}$) & is the rank of tensor $\mathcal{A}$\\
$\phi(.)$  &  denotes the feature mapping\\
$\kappa(.,.)$ & represents a kernel function\\
\bottomrule
\end{tabular}
}\vspace{-10pt}
\end{table}

\begin{defn}[Tensor product] 
The tensor product $\mathcal {A} \otimes \mathcal {B}$ of tensors $\mathcal{A} \in \mathbb{R}^{I_{1} \times I_{2} \times \cdots \times I_{N}}$ and $\mathcal{B}\in \mathbb{R}^{I_{1}' \times I_{2}' \times \cdots \times I_{M}'}$ is defined by
\begin{equation}\label{eq3}
\left(\mathcal {A} \otimes \mathcal {B}\right)_{i_1,i_2,\ldots,i_N, i_1',i_2',\ldots,i_M'}\ =\ a_{i_1,i_2,\cdots,i_N} b_{i_1',i_2',\cdots,i_M'}
\end{equation}
for all values of the indices.
\end{defn}
It is worth mentioning that a rank-one tensor, is still analogously to the matrix case, a tensor that is a tensor product of vectors ($N$th-order tensor requires $N$ vectors). Additionally, notice that for rank-one tensors $\mathcal{A}=\mathbf{a}^{(1)} \otimes \mathbf{a}^{(2)} \otimes \cdots \otimes \mathbf{a}^{(N)}$ and $\mathcal{B}=\mathbf{b}^{(1)} \otimes \mathbf{b}^{(2)} \otimes  \cdots \otimes \mathbf{b}^{(N)}$, it holds that 
\begin{equation}\label{eq4}
\left\langle \mathcal {A}, \mathcal {B}\right\rangle=\left\langle \mathbf{a}^{(1)}, \mathbf{b}^{(1)}\right\rangle \left\langle \mathbf{a}^{(2)}, \mathbf{b}^{(2)}\right\rangle \cdots \left\langle \mathbf{a}^{(N)}, \mathbf{b}^{(N)}\right\rangle.
\end{equation}

\begin{defn}[Tensor rank] The rank of a tensor $\mathcal{A}$ is the minimum number of rank-one tensor to fit $\mathcal{A}$ exactly.
\end{defn}

\begin{figure}[t]
\centering
    \label{fig_2}
    \begin{minipage}{1\columnwidth}
        \centering
        \includegraphics[width=8.0cm]{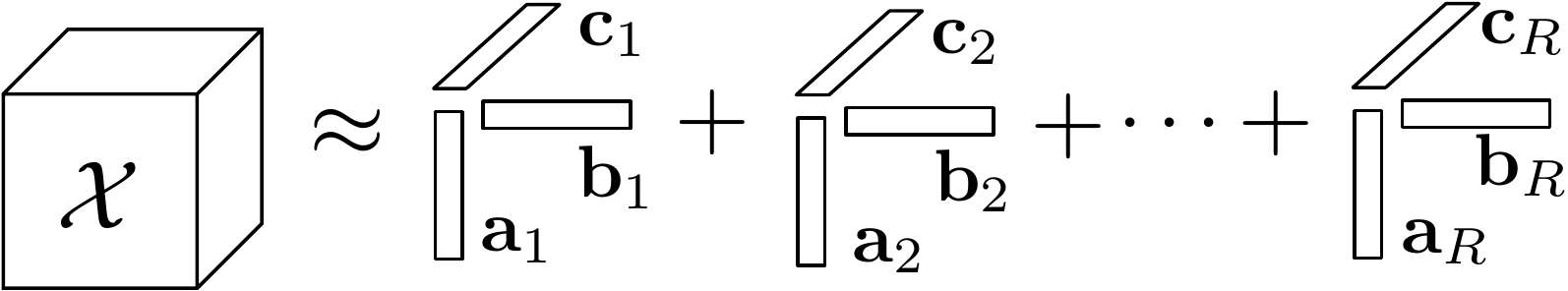}
    \end{minipage}
\caption{ CP factorization of a third-order tensor}
\label{fig-2}\vspace{-10pt}
\end{figure}

\begin{defn}[CP factorization] Given a tensor $\mathcal{A}\in \mathbb{R}^{I_{1} \times I_{2} \times \cdots \times I_{N}}$ and an integer $R$, if it can be expressed as
\begin{equation}\label{eq5}
\mathcal{A}= \sum_{r=1}^{R}\mathbf{a}_{r}^{(1)}\otimes\mathbf{a}_{r}^{(2)}\otimes \cdots \otimes \mathbf{a}_{r}^{(N)},
\end{equation}
we call it CP factorization (see Figure \ref{fig-2} for graphical representations). For convenience, in the following we write $\prod_{n=1}^{N}\otimes \mathbf{a}^{(n)}$ for $\mathbf{a}^{(1)} \otimes \mathbf{a}^{(2)} \otimes \cdots \otimes \mathbf{a}^{(N)}$.
\end{defn}

\section{APPROACH}\label{sec_method}
In this section, we first formulate the problem of tensor-based kernel learning and then elaborate on our {\ours}. For the sake of brevity, hereafter we restrict our discussion to classification problems.
\subsection{Problem statement}
Considering a training set of $M$ pairs of samples $\{\mathcal{X}_i,y_i\}_{i=1}^{M}$ for binary tensor classification problem, where $\mathcal{X}_{i} \in \mathbb{R}^{I_{1} \times I_{2} \times \cdots \times I_{N}}$ are the input of the sample and $y_i \in \{-1,+1\}$ are the corresponding class labels of $\mathcal{X}_i$. In \cite{HHCY13}, it was noted that the problem of tensor classification can be stated as a convex quadratic optimization problem in the framework of the standard linear SVM. Based on this result, we show how it can be modeled as a kernel learning problem.

Suppose we are given the optimization problem of linear tensor classification as
\begin{align}
\min_{\mathcal{W},b,\xi} & \frac{1}{2}\left\|\mathcal {W}\right\|_{F}^{2}+C \sum_{i=1}^{M} \xi_{i},\label{eq6}\\
\text{s.t.  } & y_{i} \left( \langle \mathcal{W},\mathcal{X}_{i} \rangle + b \right) \geq 1 - \xi_{i},\label{eq7}\\
& \xi_{i} \geq 0, \forall i=1,\cdots,M.\label{eq8}
\end{align}
Where $\mathcal{W}$ is the weight tensor of the separating hyperplane, $b$ is the bias, $\xi_{i}$ is the error of the $i$th training sample, and $C$ is the trade-off between the classification margin and misclassification error.

Obviously, the optimization problem in (\ref{eq6})-(\ref{eq8}) is the generalization of the problem of the standard linear SVM to tensor patterns in tensor space. When the input samples $\mathcal{X}_i$ are vectors, it degenerates into the standard linear SVM. As such, based on the kernel method for the extension of linear SVM to the nonlinear case$-$by introducing a nonlinear feature mapping $\phi: \mathbf{x} \rightarrow \phi \left(\mathbf{x}\right)\in \mathfrak{H} \subset \mathbb{R}^{H}$, we develop a nonlinear extension of (\ref{eq6})-(\ref{eq8}) in the following, which is critical for the derivation of the model for tensor-based kernel learning.
\indent Given a tensor $\mathcal{X} \in \mathbb{R}^{I_{1} \times I_{2} \times \cdots \times I_{N}}$, we assume it is mapped into the Hilbert space $\mathfrak{H}$ by
\begin{equation}\label{eq9}
\phi: \mathcal{X} \rightarrow \phi \left(\mathcal{X}\right) \in \mathbb{R}^{H_{1} \times H_{2} \times \cdots \times H_{P}}.
\end{equation}
\noindent Note that the project tensor $\phi\left(\mathcal{X}\right)$ in space $\mathfrak{H}$ may have different order with $\mathcal{X}$, and each mode dimension is higher even an infinite dimension depending on the feature mapping function $\phi(.)$. Such a Hilbert space is called the high-dimensional tensor feature space or simply a tensor feature space. According to the same principle as the construction of linear classification model in the original tensor space, we construct the following model in this space:
\begin{align}
\min_{\mathcal{W},b,\xi} & \frac{1}{2}\left\|\mathcal {W}\right\|_{F}^{2}+C \sum_{i=1}^{M} \xi_{i},\label{eq10}\\
\text{s.t.  } & y_{i} \left( \langle \mathcal{W}, \phi\left(\mathcal{X}_{i}\right) \rangle + b \right) \geq 1 - \xi_{i},\label{eq11}\\
& \xi_{i} \geq 0, \forall i=1,\cdots,M.\label{eq12}
\end{align}
\noindent From the viewpoint of high-dimensional tensor feature space, this model is a linear model. However, from the viewpoint of the original tensor space, it is a nonlinear model. When the input samples $\mathcal{X}_{i}$ are vectors, it degenerates into the standard nonlinear SVM. When the feature mapping function $\phi(.)$ is an identical function, {\ie}, $\phi(\mathcal{X})=\mathcal{X}$, it is the same as that in (\ref{eq6})-(\ref{eq8}). Thus, we say that the optimization model (\ref{eq10})-(\ref{eq12}) is the nonlinear counterpart of (\ref{eq6})-(\ref{eq8}).

Let us now show how this model can be exploited to obtain tensor-based kernel optimization model. Using Lagrangian relaxation method \cite{ES01}, it is easy to check that the dual problem of (\ref{eq10})-(\ref{eq12}) is
\begin{align}
\max_{\alpha_1, \alpha_2, \cdots, \alpha_M} & \sum_{i=1}^{M}\alpha_i - \frac{1}{2}\sum_{i, j=1}^{M}\alpha_i\alpha_j y_i y_j \langle \phi\left(\mathcal{X}_{i}\right), \phi\left(\mathcal{X}_{j}\right) \rangle \label{eq13}\\
\text{s.t.} &  \sum_{i=1}^M \alpha_i y_i=0, \label{eq14}\\
& 0\leq\alpha_i\leq C, \forall i=1,\cdots,M. \label{eq15}
\end{align}
Where $\alpha_i$ are the Lagrangian multipliers and $\langle \phi\left(\mathcal{X}_{i}\right), \phi\left(\mathcal{X}_{j}\right) \rangle$ are the inner product between the mapped tensors of $\mathcal{X}_i$ and $\mathcal{X}_j$ in the tensor feature space.

The advantage of formulation (\ref{eq13})-(\ref{eq15}) over (\ref{eq10})-(\ref{eq12}) is that the training data only appear in the form of inner products. Based on the fundamental principle of kernel method, by substituting the inner product $\langle \phi\left(\mathcal{X}_{i}\right), \phi\left(\mathcal{X}_{j}\right) \rangle$ with a suitable tensor kernel function $\kappa\left(\mathcal{X}_i,\mathcal{X}_j\right)$, we thus get the tensor-based kernel model. The resulting decision function is
\begin{equation} \label{eq19}
f \left (\mathcal{X}\right)=sign\left(\sum_{i=1}^{M}\alpha_i y_i \kappa\left(\mathcal{X}_i,\mathcal{X}\right) +b\right).
\end{equation}

\subsection{{\ours}}
From the above statement, we can see that tensor-based kernel learning degenerates into the study of kernel function, and the success of kernel methods depends strongly on the data representation encoded into the kernel function. Now we propose the {\ours}. Our target is to leverage the naturally available structure of the tensor to facilitate kernel learning.

Tensors provide a natural and efficient representation for multi-way data, but there is no guarantee that such representation will be good for kernel learning. Since learning will only be successful if the regularities that underlie the data can be discerned by the kernel. As with the previous analysis for the characteristics of tensor object, we know that the essential information in the tensor is embedded in its multi-way structure. Thus, one important aspect of kernel learning for such complex objects is to represent them by sets of key structural features easier to manipulate, and design kernels on such sets. 

According to the mathematical definition of tensor, we can gain a further understanding of the structure of the tensor that tensor object has the tensor product structure. In previous work, it was found that CP factorization is particularly effective for extracting this structure. Motivated by these observations, we investigate how to exploit the benefits of CP factorization to learn a structure-preserving kernel in the tensor product feature space. More specifically, we will represent each tensor object as a sum of rank-one tensors in the original space and map them into the tensor product feature space for our kernel learning. In the following, we illustrate how to design the feature mapping.

We start by defining the following mapping on a rank-one tensor $\prod_{n=1}^{N}\otimes \mathbf{x}^{(n)} \in \mathbb{R}^{I_{1} \times I_{2} \times \cdots \times I_{N}}$.
\begin{equation}\label{eq20}
\phi: \prod_{n=1}^{N}\otimes \mathbf{x}^{(n)} \rightarrow \prod_{n=1}^{N}\otimes \phi \left( \mathbf{x}^{(n)} \right) \in \mathbb{R}^{H_{1} \times H_{2} \times \cdots \times H_{N}}.
\end{equation}
\begin{figure}[t]
\centering
        \centering
        \includegraphics[width=8 cm]{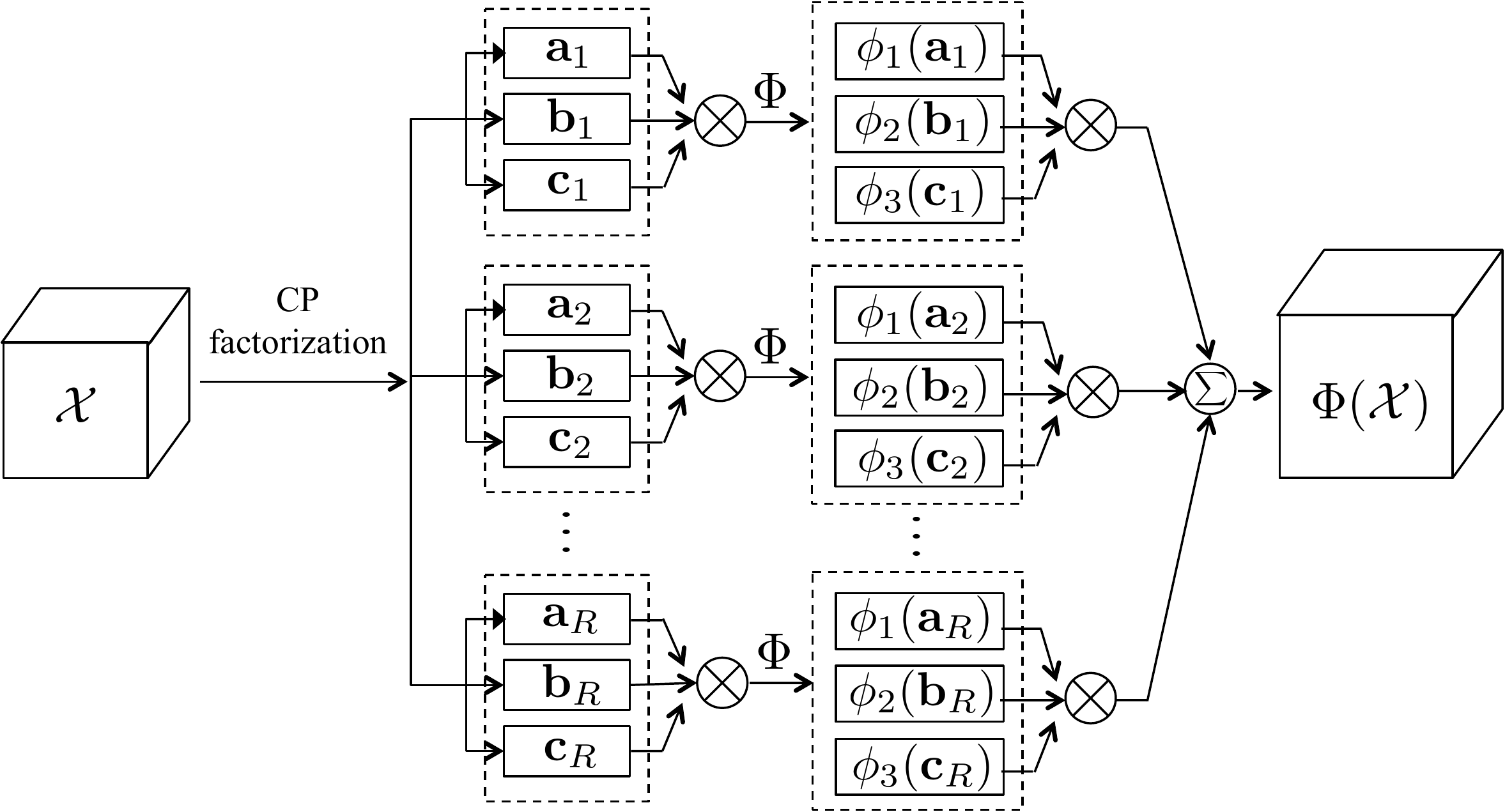}\vspace{-8pt}
\caption{ Dual-tensorial mapping}
\label{fig3}\vspace{-10pt}
\end{figure}

\noindent Let the CP factorization of $\mathcal{X}, \mathcal{Y} \in \mathbb{R}^{I_{1} \times I_{2} \times \cdots \times I_{N}}$ be $\mathcal{X}= \sum_{r=1}^{R} \prod_{n=1}^{N}\otimes \mathbf{x}_{r}^{(n)}$ and $\mathcal{Y}= \sum_{r=1}^{R} \prod_{n=1}^{N}\otimes \mathbf{y}_{r}^{(n)}$ respectively. By using the concept of the kernel function, we see that the kernel can be defined directly with inner product in the feature space. Thus, when $R=1$, based on the above mapping and Eq.~\ref{eq4}, we can directly derive the naive tensor product kernels, {\ie},
\begin{align}
 & \kappa\left(\mathcal{X},\mathcal{Y}\right) =\prod_{n=1}^{N}\kappa \left(\mathbf{x}^{(n)},\mathbf{y}^{(n)}\right).\label{eq21}
\end{align}
Despite this, many authors has demonstrated that a simple rank-one tensor cannot provide compact and informative presentation for original data \cite{ZHL12}. The key point is how to design feature mapping when the value of $R$ is more than one.

Based on the definition of the kernel function, it is easy to find that the feature space is a high-dimensional space of the original space, equipped with the same operations. Thus, we can factorize tensor data directly in the feature space the same as original space. This is formally equivalent to performing the following mapping:
\begin{equation}\label{eq22}
\phi: \sum_{r=1}^{R}\prod_{n=1}^{N}\otimes \mathbf{x}_{r}^{(n)} \rightarrow \sum_{r=1}^{R} \prod_{n=1}^{N}\otimes \phi \left( \mathbf{x}_{r}^{(n)} \right).
\end{equation}

\noindent In this sense, it corresponds to mapping tensors into high-dimensional tensors that retain the original structure. More precisely, it can be regarded as mapping the original data into tensor feature space and then conducting the CP factorization in the feature space. We call it the dual-tensorial mapping function (see Figure~\ref{fig3}).

After mapping the CP factorization of the data into the tensor product feature space, the kernel itself is just the standard inner product of tensors in that feature space. Thus, we derive our {\ours}:
\begin{equation}
\label{eq23}
\begin{split}
\kappa \left( \sum_{r=1}^{R} \prod_{n=1}^{N} \otimes \mathbf{x}_{r}^{(n)}\ , \ \sum_{r=1}^{R} \prod_{n=1}^{N} \otimes \mathbf{y}_{r}^{(n)} \right) \\
= \sum_{i=1}^{R} \sum_{j=1}^{R} \prod_{n=1}^{N} \kappa \left( \mathbf{x}_{i}^{(n)},\mathbf{y}_{j}^{(n)}\right)
\end{split}
\end{equation}

\noindent From its derivation, we know such a kernel can take the multi-way structure flexibility into account. In general, the {\ours} is an extension of the conventional kernels in the vector space to tensor space, and each vector kernel can be used in this framework for supervised tensor learning in conjunction with kernel machines.

\subsection{Efficiency}
We consider the case of Gaussian RBF kernel in our framework, which is one of the most popular kernels that have been proven successful in many different contexts. Assume that a set of tensor data $\left\{(\mathcal{X}_i ,\ y_i)\right\}_{i=1}^{M}$ is given, where $\mathcal{X}_i\in \mathbb{R}^{I_{1} \times I_{2} \times \cdots \times I_{N}}$. The time complexity of computing a Gaussian RBF kernel matrix is $O\left(M^2\prod_{n=1}^{N}I_n\right)$ and our method {\ours} is thus $O\left(M^2R^2\sum_{n=1}^{N}I_n\right)$. A typical characteristic associated with tensor data is very high dimensional while $R$ is often very small, which indicates our proposed method is significantly more efficient than its vector counterpart. It is also worth mentioning that our method depends on CP factorization technique, but it is backed with rapid implementation \cite{LL08}. The storage complexity is reduced from $O\left(M\prod_{n=1}^{N}I_n\right)$ to $O\left(M\sum_{n=1}^{N}I_n\right)$, where the data is compressed without quality loss and can be recovered quickly.
Furthermore, since the constituent kernels are Gaussian RBF kernels, we can thus reformulate Eq.~\ref{eq23} to
\begin{equation}
\label{eq24}
\begin{split}
 \kappa\left(\mathcal{X},\mathcal{Y}\right)=\sum_{i=1}^{R}\sum_{j=1}^{R}\prod_{n=1}^{N}\kappa\left(\mathbf{x}_{i}^{(n)},\mathbf{y}_{j}^{(n)}\right)\\
= \ \sum_{i=1}^{R}\sum_{j=1}^{R}\exp\left(-\sigma\sum_{n=1}^{N}\|\mathbf{x}_{i}^{(n)}-\mathbf{y}_{j}^{(n)}\|^2\right)
\end{split}
\end{equation}
where $\sigma$ is used to set an appropriate bandwidth. We denote this kernel as {\ourmethod}.

\section{Experiment Evaluation} \label{sec_exp}
In this study, we validate the effectiveness of the {\ourmethod} kernel within standard SVM framework for tensor classification, which we refer to as {\ourmethod}. As an application we consider an example of neuroimaging mining.

\subsection{Data collection}
We use three real-world fMRI datasets in our experimental evaluation as follows.

\begin{figure}[t]
\centering
\subfigure[]
{
    \label{fig_example1_a}
    \begin{minipage}{.4\columnwidth}
        \centering
        \includegraphics[width=4.0cm]{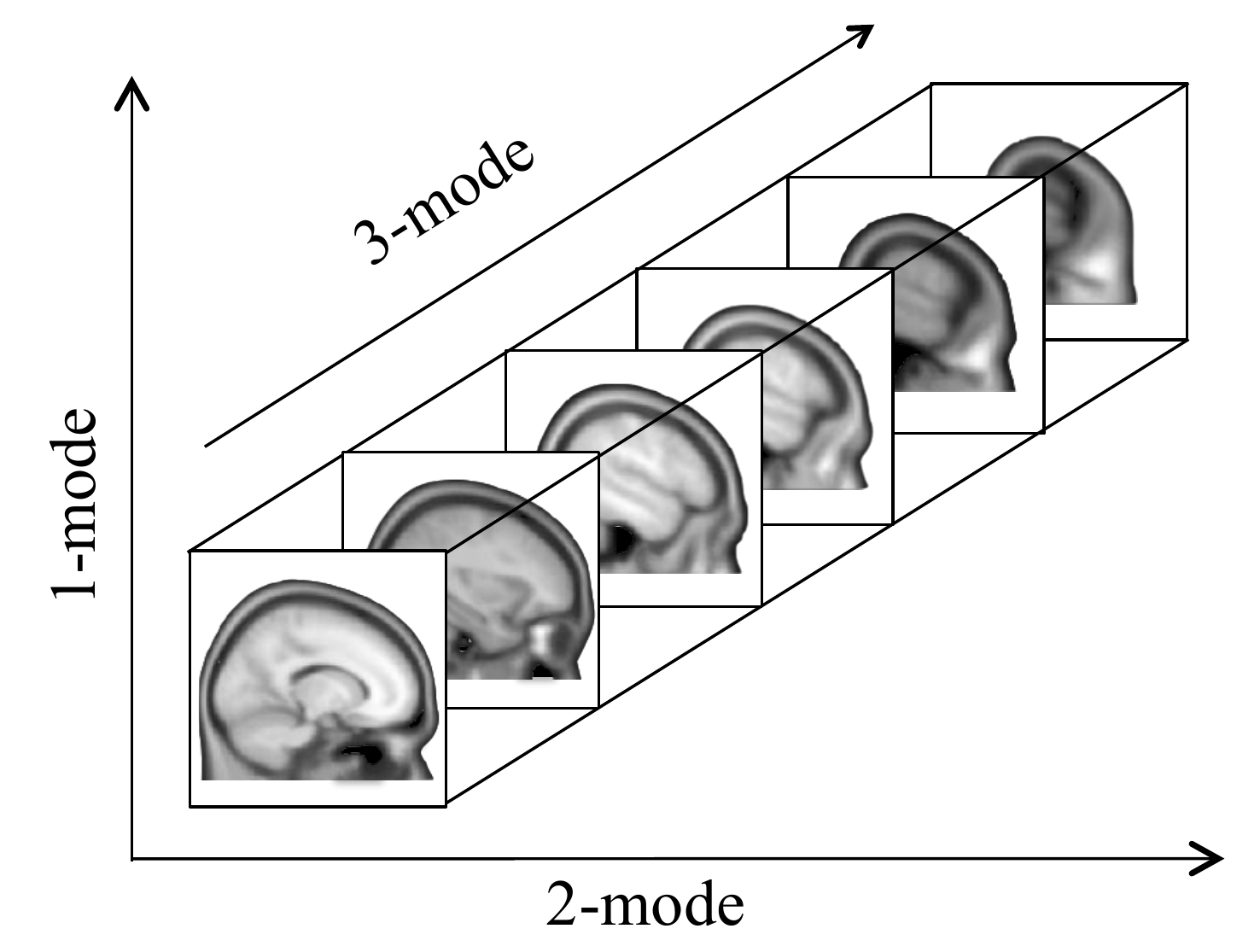}
    \end{minipage}
}
\subfigure[]
{
    \label{fig_example1_b}
    \begin{minipage}{.4\columnwidth}
        \centering
        \includegraphics[width=2cm]{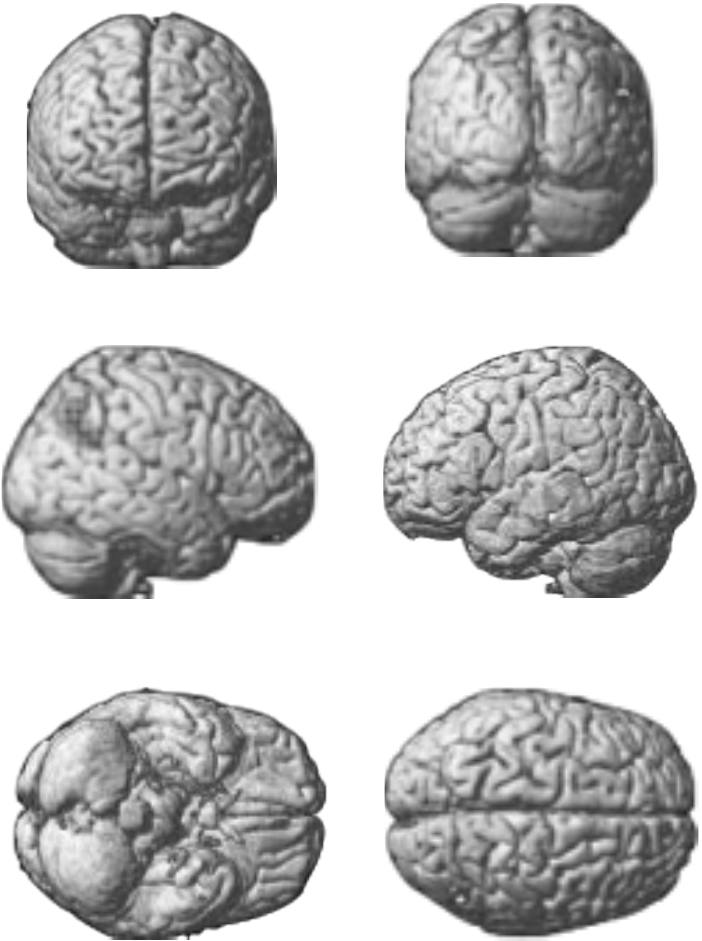}
    \end{minipage}
}\vspace{-8pt}
\caption{(a) An illustration of a three-order tensor (fMRI image), (b) An visualization of fMRI image.}
\label{fig4}\vspace{-10pt}
\end{figure}

\noindent\textbullet \ 
\emph{Alzheimer's Disease (ADNI)}: The first dataset is collected from the Alzheimer's Disease Neuroimaging Initiative\footnote{\url{http://adni.loni.usc.edu/}}. The dataset consists of records of patients with Alzheimer's Disease (AD) and Mild Cognitive Impairment (MCI). We downloaded all records of resting-state fMRI images and apply SPM8 toolbox\footnote{\url{http://www.ﬁl.ion.ucl.ac.uk/spm/software/spm8/}} to preprocess the data. We deleted the first ten volumes for each individual, and functional images were realigned to the first volume, slice timing corrected, and normalized to the MNI template and spatially smoothed with an 8-mm FWHM Gaussian kernel. Resting-State fMRI Data Analysis Toolkit (REST\footnote{\url{http://resting-fmri.sourceforge.net}}) was then used to remove the linear trend of time series and temporally band-pass filtering ($0.01-0.08$ Hz). The average value of each subject over time series was calculated within each of those boxes, thereby resulting in 33 samples and a sum total of $61\times73\times61=271633$ voxels (or features). We treat the normal brains as negative class, and AD+MCI as the positive class. Each individual is linearly rescaled to $[0, 1]$. Feature normalization is an important procedure, since the brain of every individual is different.\\
\noindent\textbullet \ 
\emph{Attention Deficit Hyperactivity Disorder (ADHD)}: The second dataset is collected from ADHD-200 global competition dataset\footnote{\url{http://neurobureau.projects.nitrc.org/ADHD200/}}. The dataset contains records of resting-state fMRI images for $776$ subjects with $58\times49\times47=133574$ voxels, which are labeled as real patients (positive) and normal controls (negative). The original dataset is unbalanced, we randomly sampled $100$ ADHD patients and $100$ normal controls from the dataset for performance evaluation and the average over time series is conducted. Such dataset are quite special, all algorithms perform bad with normalization, we use non-normalized dataset.\\
\noindent\textbullet \ 
\emph{Human Immunodeficiency Virus Infection (HIV)}: The third dataset is collected from the Department of Radiology in Northwestern University \cite{WF11}. The dataset contains fMRI brain images of patients with early HIV infection (positive) as well as normal controls (negative). The same preprocessing steps as in ADNI dataset were given. This contains $83$ samples with $61\times73\times61=271633$ voxels.

\subsection{Baselines and Metrics}
In order to establish a comparative study, we use seven state-of-the-art methods as baselines, each representing a different strategy. We here focus on SVM classifier, since it has been proven successful in many applications.

\begin{table*}[t]
{\tiny
\centering
    \caption{Average classification accuracy comparison: mean (standard deviation).}\label{tb_2}
\begin{tabular}{|l|c|c|c|c|c|c|c|c|}
\hline
    \textbf{Dataset}
    & \textbf{\ourmethod}
    & \textbf{Gaussian RBF}
    & \textbf{Factor kernel}
    & \textbf{K$_{3rd}$ kernel}
    & \textbf{linear SHTM}
    & \textbf{linear SVM}
    & \textbf{PCA+SVM}
    & \textbf{MPCA+SVM}\\
\hline
ADNI    & \textbf{0.75 (0.18)}   & 0.49 (0.23)   & 0.51 (0.21)   & 0.55 (0.14)   & 0.52 (0.31)   & 0.42 (0.27)    & 0.50 (0.02)   & 0.51 (0.02)\\
ADHD    & \textbf{0.65(0.01)}    & 0.58 (0.00)   &      0.50 (0.00)       & 0.55 (0.00)   & 0.51 (0.03)  & 0.51 (0.01)    & 0.63 (0.01)   & 0.64 (0.01)\\
HIV     & \textbf{0.74 (0.00)}   & 0.70 (0.00)   & 0.70 (0.01)   & \textbf{0.75 (0.02)}   & 0.70 (0.01)  & 0.74 (0.01)    & 0.73 (0.25)   & 0.72 (0.02)\\
\hline
\end{tabular}
}\vspace{-10pt}
\end{table*}
\begin{figure*}[t]
\centering
\subfigure[ADNI]
{
    \label{fig_5_a}
    \begin{minipage}{.4\columnwidth}
        \centering
        \includegraphics[width=\textwidth]{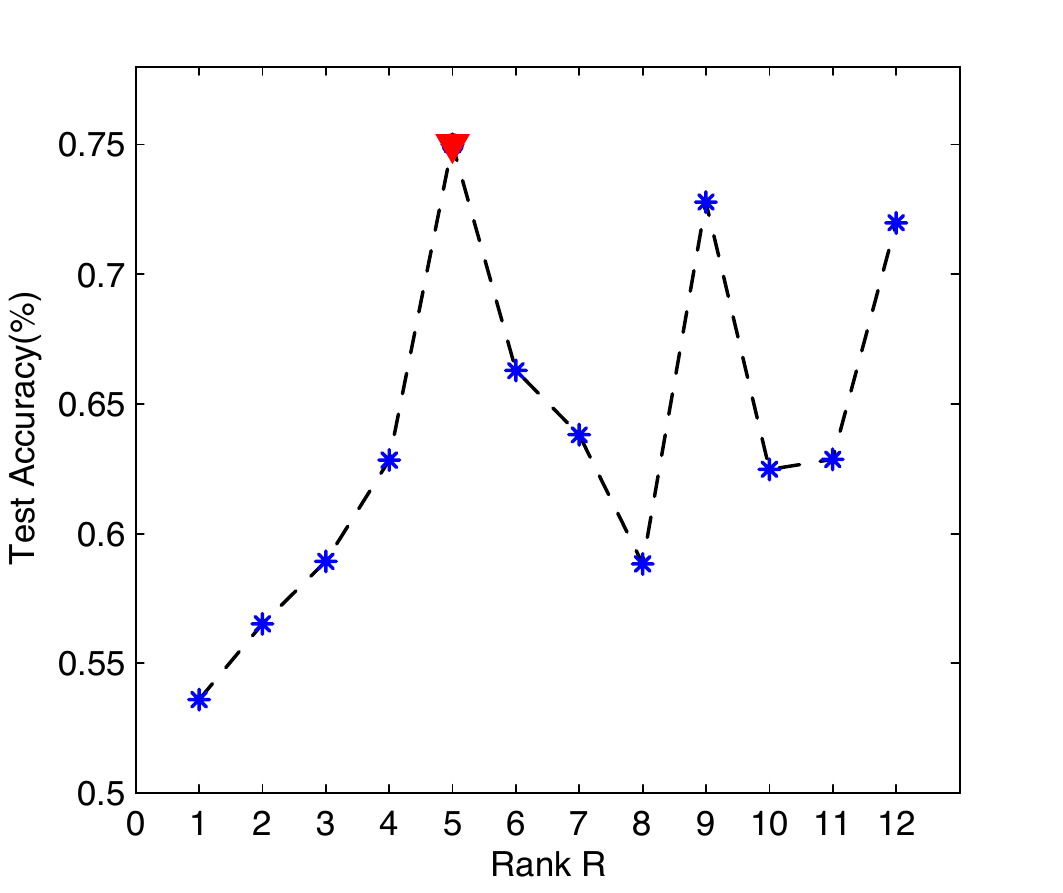}
    \end{minipage}
}
\subfigure[ADHD]
{
    \label{fig_5_b}
    \begin{minipage}{.4\columnwidth}
        \centering
        \includegraphics[width=\textwidth]{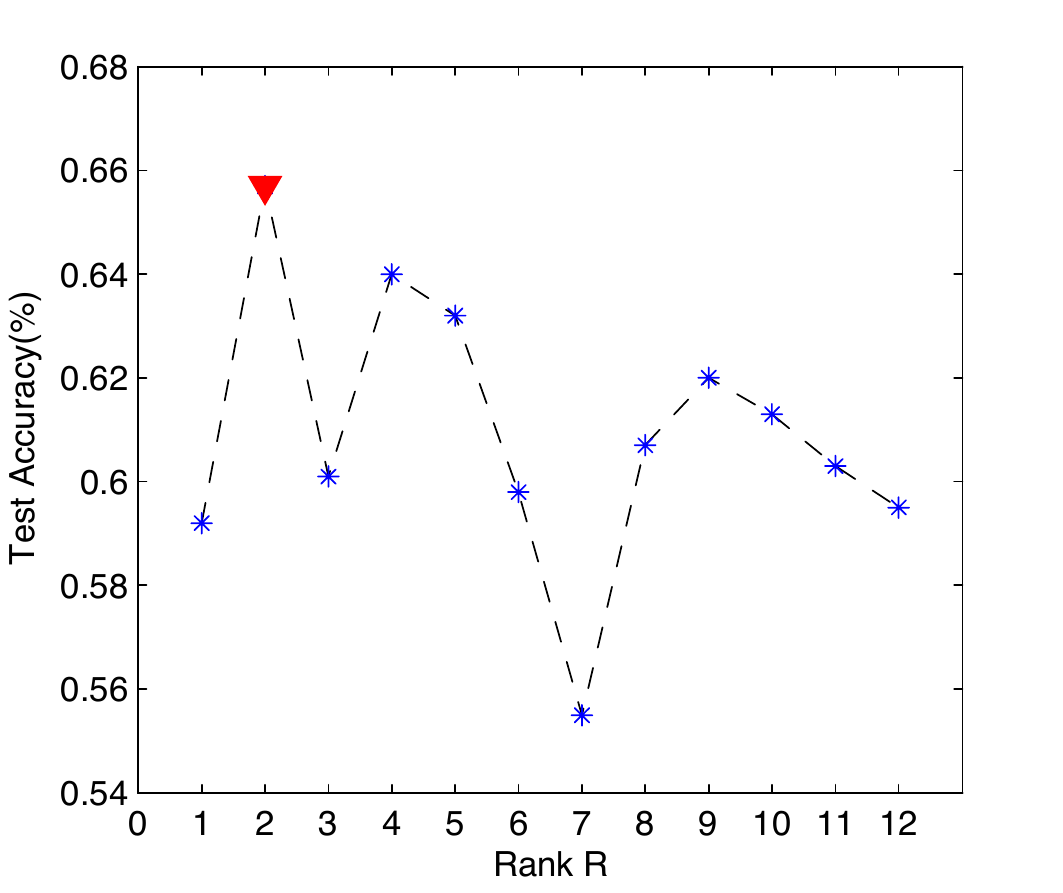}
    \end{minipage}
}
\subfigure[HIV]
{
    \label{fig_5_c}
    \begin{minipage}{.4\columnwidth}
        \centering
        \includegraphics[width=\textwidth]{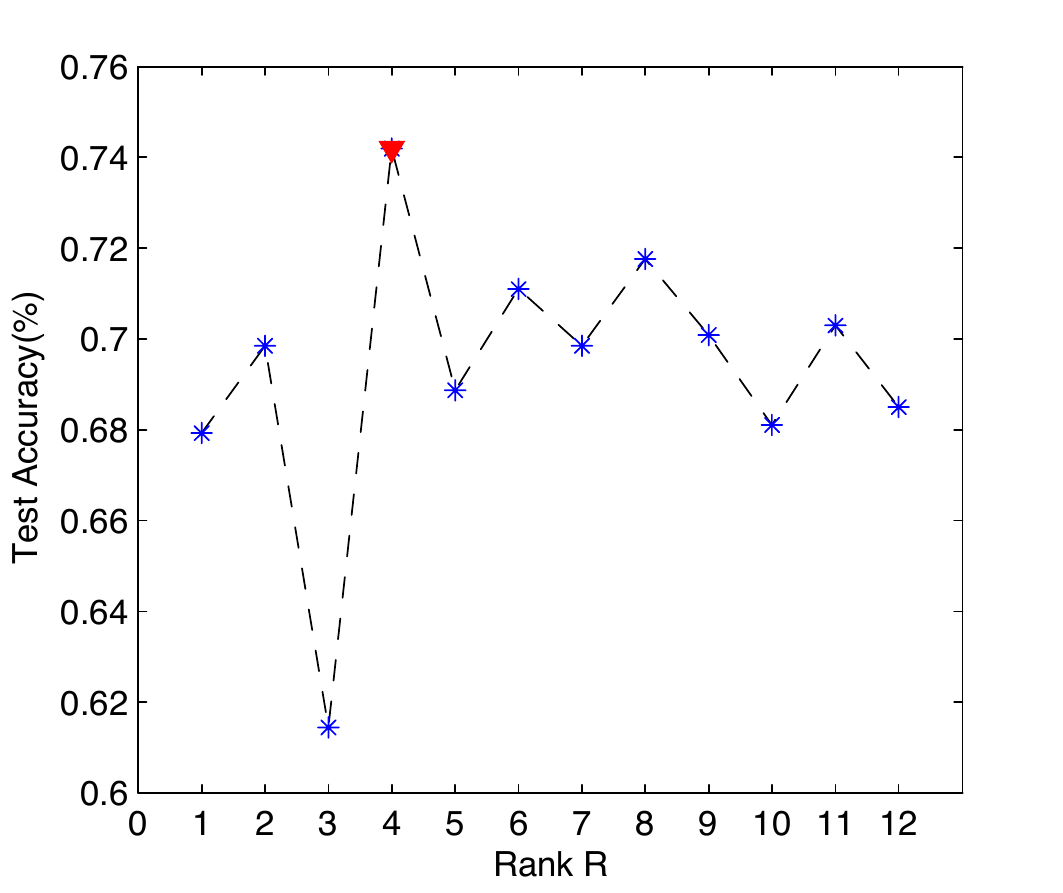}
    \end{minipage}
    
}\vspace{-8pt}
\caption{Test accuracy vs. $R$ on (a) ADNI, (b) ADHD, and (c) HIV, where the red triangles indicate the peak positions.}
\label{fig-5}\vspace{-10pt}
\end{figure*}

\noindent\textbullet \ Gaussian-RBF: a Gaussian-RBF kernel-based SVM, which is now the most widely used vector-based method for classification. In the following methods, if not stated explicitly, we use SVM with Gaussian RBF kernel as the classifier.\\
\noindent\textbullet \ 
Factor kernel: a matrix unfolding based tensor kernel, which is recently proposed in \cite{SLS11} and the constituent kernels belong to a class of Gaussian RBF kernels.\\
\noindent\textbullet \ 
K$_{3rd}$ kernel: a class of vector-based tensor kernels, aiming at representing the tensor in each vector space to capture structural information and have been applied to analyze fMRI images in conjunction with Gaussian RBF kernel \cite{S11}.
\\
\noindent\textbullet \ 
Linear SHTM: a linear support higher-order tensor machine \cite{HHCY13}, which is one of the most effective methods for tensor classification that generalizes linear SVM to tensor pattern using CP factorization and can be regarded as a special case of {\ours}, namely the constituent kernels are linear kernels. This baseline is used to test the ability of our proposed method to cope with complex (possibly nonlinear) structured data.
\\
\noindent\textbullet \ 
Linear kernel: linear SVM has also been increasingly used to handle fMRI data. In some cases, it outperforms SVM using nonlinear kernels.
\\
\noindent\textbullet \ 
PCA+SVM: Principal component analysis (PCA) is a vector-based subspace learning algorithms, which are commonly used for dealing with high-dimensional data, in particular fMRI data.
\\
\noindent\textbullet \ 
MPCA+SVM: Multilinear principal component analysis (MPCA) \cite{LPV08} is a natural extension of PCA to tensors, which are used to handle high-dimensional tensor data.

The first three baselines are used to show the improvement of our proposed method over current kernel approaches to tensor classification.
The last two baselines are used to test the effectiveness of our proposed method compared to unsupervised methods for tensor classification.

The effectiveness of an algorithm is always evaluated by test accuracy, we utilize it as metrics in the experiments. For our proposed method and linear SHTM, we choose the most popular and widely used enhanced linear search method \cite{LL08} as its CP factorization strategy. All of the related methods select the optimal trade-off parameter from $C\in \{2^{-5}, 2^{-4}, \cdots,2^{9}\}$ and kernel width parameter from $\sigma\in \{2^{-4}, 2^{-3}, \cdots, 2^{9}\}$. Considering the fact that there is no known closed-form solution to determine the rank $R$ of a tensor a priori \cite{KM11}, and rank determination of a tensor is still an open problem \cite{SL08}, in our method and linear SHTM, we use grid search to determine the optimal rank and the optimal trade-off parameter together, where the rank $R\in \{1, 2, \cdots, 12\}$. The influence of different rank parameters on the classification performance of our method is also given.

All the experiments are conducted on a computer with Intel Core2\texttrademark 1.8GHz processor and 3.5GB RAM memory running Microsoft Windows XP.

\subsection{Classification Performance}
In our experiments, we first randomly sample 80\% of the whole data as the training set, and the remaining samples as the test set. This random sampling experiment was repeated 50 times for all methods. The average performances of each method are reported. Table~\ref{tb_2} shows the average classification accuracy and standard deviation of seven algorithms on three datasets, where the best result is highlighted in bold type.

From the experimental results in Table~\ref{tb_2}, we can observe that the classification accuracy of each method on different dataset can be quite different. However, the best method that outperforms other methods in all datasets is {\ourmethod}, especially for ADNI dataset. It is worth noting that in neuroimaging task it is very hard for classification algorithms to achieve even moderate classification accuracy on ADNI dataset since this data is extremely high dimensional but with small sample size. While we can observe an 20\% gain over comparison methods. Based on this result, we can conclude that operation on tensors is much more effective than on matrices and vectors for high-dimensional tensor data analysis.

\begin{figure}[t]
\centering
\subfigure[original data]
{
    \label{fig6a}
    \begin{minipage}{.35\columnwidth}
        \centering
        \includegraphics[width=\textwidth]{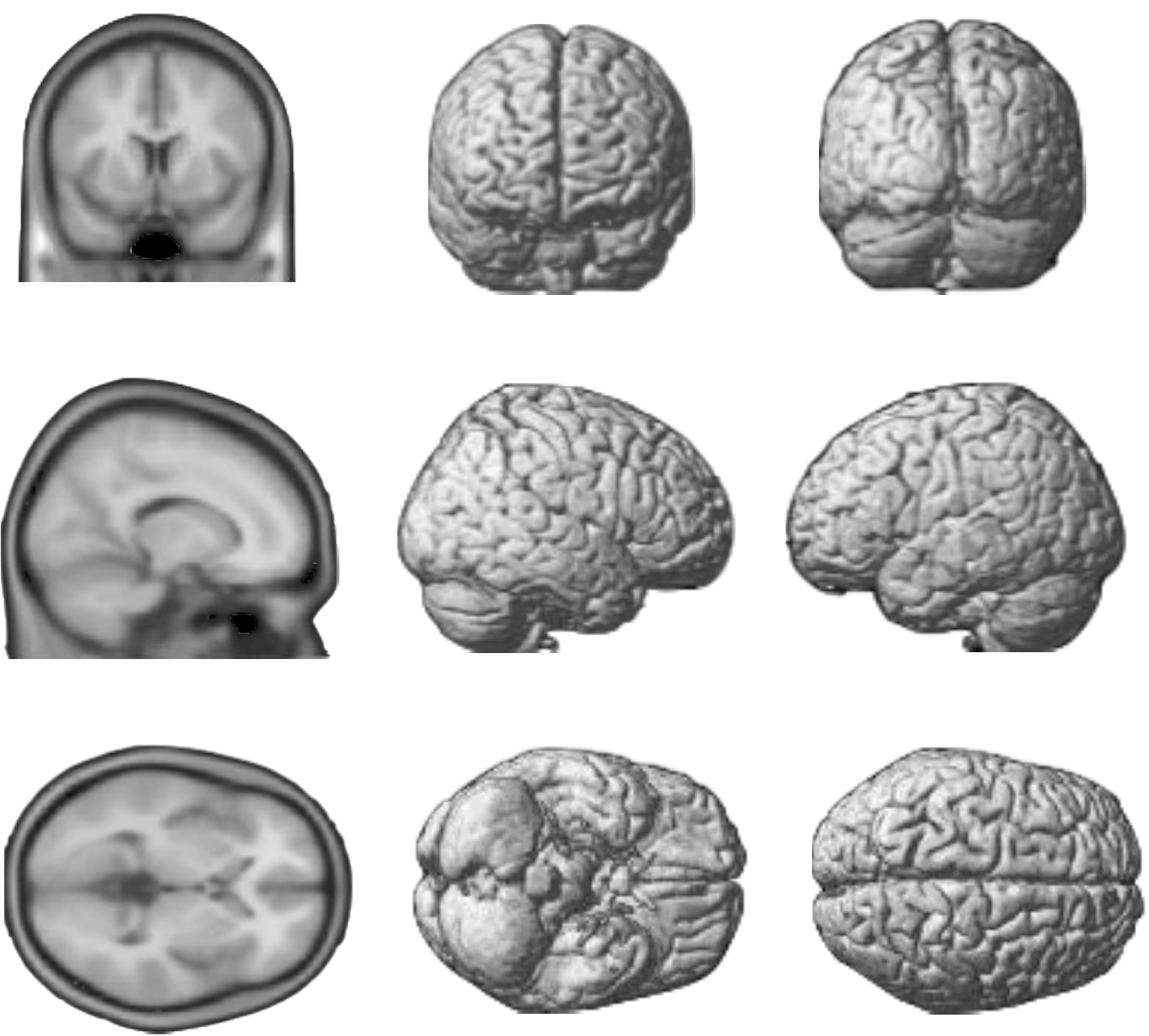}
    \end{minipage}
}
\subfigure[reconstruction]
{
    \label{fig6b}
    \begin{minipage}{.35\columnwidth}
        \centering
        \includegraphics[width=\textwidth]{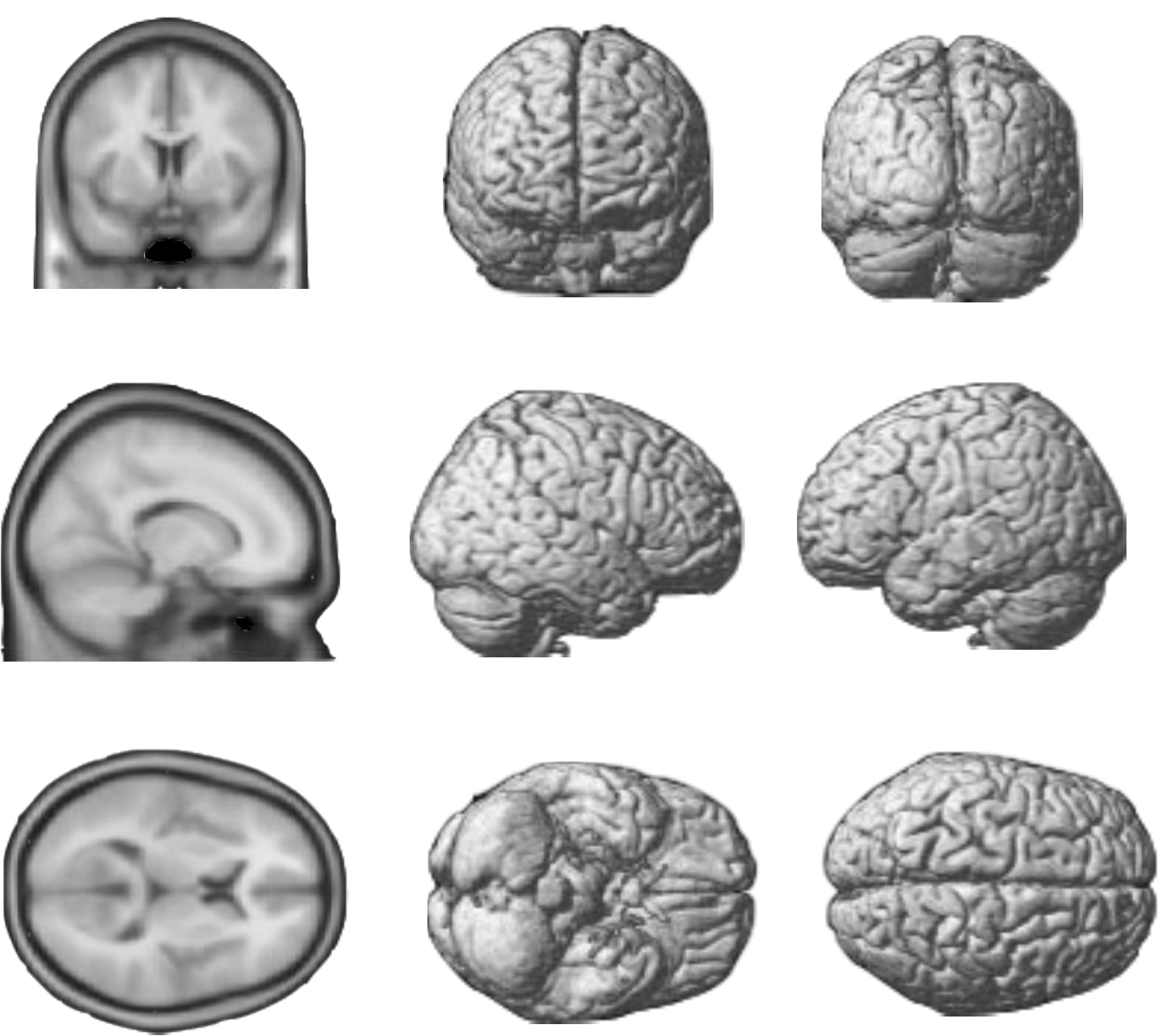}
    \end{minipage}
}\vspace{-8pt}
\caption{(a) is visualization of original ADNI object (a cross section is shown on the left and a 3D plot on the right) and (b) is reconstruction result from our chosen CP factorization.}
\label{fig-6}\vspace{-10pt}
\end{figure}

So far we have demonstrated that our proposed method is effective for tensor classification. However, it is still interesting to show how the data structure for tensor is actually used in our method. We focus on ADNI dataset to conduct an analysis. Figure~\ref{fig-6} shows the visualization of original ADNI object and reconstruction result from our chosen CP factorization. As illustrated, CP factorization can fully capture the multi-way structure of the data, thus our method take it into account in the learning process.

\subsection{Parameter Sensitivity}
Although the optimal rank parameter $R$ , the optimal trade-off parameter $C$ and kernel width parameter $\sigma$ are found by a grid search in {\ourmethod}, it is still important to see the sensitivity of {\ourmethod} to the rank parameter $R$. For this purpose, we demonstrate a sensitivity study over different $R\in\{1, 2, \cdots, 12\}$ in this section, where the optimal trade-off parameter and kernel width parameter are still selected from $C\in \{2^{-5}, 2^{-4}, \cdots,2^{9}\}$ and $\sigma\in \{2^{-4}, 2^{-3}, \cdots, 2^{9}\}$ respectively. According to the aforementioned analysis, we know that the efficiency of {\ourmethod} is reduced when $R$ is increased because a higher value of $R$ implies that more items are included into kernel computations. Thus, we only demonstrate the variation in test accuracy over different $R$ on three datasets. As shown in Figure~\ref{fig-5}, we can observe that the rank parameter $R$ has a significant effect on the test accuracy and the optimal value of $R$ depends on the data, while the optimal value of $R$ lies in the range $2 \leq R\leq 5$, which may provide a good guidance for selection of the $R$ in advance.

In summary, the parameter sensitivity study indicates that the classification performance of {\ourmethod}+SVM relies on parameter $R$ and it is difficult to specify an optimal value for $R$ in advance. However, in most cases the optimal value of $R$ lies in a small range of values as demonstrated in \cite{HHCY13} and it is not time-consuming to find it using the grid search strategy in practical applications.

\section{Related Work}\label{sec_related}
From the conceptual perspective, two topics can be seen as closely related to our {\ours} approach: supervised tensor learning and tensor factorization. This section gives a short overview of these areas and distinguishes {\ours} from other existing solutions.

\textbf{Tensor factorizations}:
Tensor factorizations are higher-order extensions of matrix factorization that elicit intrinsic multi-way structures and capture the underlying patterns in tensor data. These techniques have been widely used in diverse disciplines to analyze and process tensor data. A thorough survey of these techniques and applications can be found in \cite{KB09}. The two most commonly factorizations are CP and Tucker. CP is a special case of Tucker decomposition which forces the core array to a (super)diagonal form. It is thus more condensed than that of Tucker. In the supervised tensor learning setting, CP is more frequently applied to explore tensor data because of its properties of uniqueness and simplicity \cite{HHCY13, AIA13,TLWH07,ZLZ12}. However, in these applications, CP factorization is used either for exploratory analysis or to deal with linear tensor-based models. In this study, we employ the CP factorization to foster the use of kernel methods for supervised tensor learning.

\textbf{Supervised tensor learning}:
Supervised tensor learning has been extensively studied in recent years \cite{CHH06,GKP12,KP11,TLWH07,ZLZ12}. Most of previous work has concentrated on learning linear tensor-based models, whereas the problem of how to build nonlinear models directly on tensor data has not been well studied. A first attempt in this direction focused on second-order tensors and led to a non-convex optimization problem \cite{SLS10}. Subsequently, the authors claimed that it can be extended to deal with higher-order tensors at the cost of a higher computational complexity, and proposed a factor kernel for tensors of arbitrary order except for square matrices based upon matrix unfoldings \cite{SLS11}. In the context of this proposal, Signorette et al. \cite{SOLS12} introduced a cumulant-based kernel approach for classification of multichannel signals. Zhao et al. \cite{ZZAZC13} presented 
a kernel tensor partial least squares for regression of lamb movements. A drawback of the approaches in \cite{SLS11,SOLS12,ZZAZC13} is that they can only capture the one-way relationships within the tensor data, because the tensors are unfolded into matrices. The multi-way structures within tensor data are already lost before the kernel construction process. Different from these methods, we aim to directly exploit the algebraic structure of the tensor to study structure-preserving kernels.

Another recent work by Hardoon et al. \cite{HS10}, although not directly performs supervised tensor learning, is worth mentioning in this context. They introduced the so-called tensor kernels to analyze neuroimaging data from multiple sources, which demonstrated that the tensor product feature space is useful for modeling interactions between feature sets in different domains. In this study, we make use of the tensor product feature space to derive our kernels in vivo the incorporation of CP model. The tensor kernels can be cast as a special case of our framework.

\section{Conclusion and Future work} \label{sec_conclusion}
In this paper we have introduced a new tensor-based kernel methodology and first operate directly on tensors. We have applied our method on the problem of fMRI classification. The results indicate that the prior structural information can indeed improve the classification performance, particularly with small-sample size. As previous work limited on learning with matrices and vectors, this paper provides a new insight into the understanding of the principles and ideas underlying the concept of tensor.

In the future, we will investigate the reconstruction techniques of tensor data, so that our method can handle high-dimensional vector data more effectively. Another interesting topic would be to design some special method to address the parameter problem. Further study on this topic will also include many applications of DuSK kernels in real-world unsupervised learning with tensor representations.
\vspace{-8pt}

\section*{Acknowledgements}
{\scriptsize
This work is supported in part by NSF through grants CNS-1115234, DBI-0960443, and OISE-1129076, NIH through grant MH080636, US Department of Army through grant W911NF-12-1-0066, Huawei Grant, National Science Foundation of China ($61273295$, $61070033$), National Social Science Foundation of China ($11$\&ZD$156$), Science and Technology Plan Project of Guangzhou City($12C42111607$, $201200000031$), Science and Technology Plan Project of Panyu District Guangzhou ($2012$-Z-$03$-$67$), Specialized Research Fund for the Doctoral Program of Higher Education ($20134420110010$). Discipline Construction and Quality Engineering of Higher Education in Guangdong Province(PT$2011$JSJ) and China Scholarship Council.

}
\vspace{-8pt}
\balance
\bibliographystyle{abbrv}
{\tiny

}
\end{document}